\documentclass[sigconf]{acmart}

\usepackage{booktabs} 
\usepackage{moreverb}
\usepackage{paralist}
\usepackage{enumitem}

\setcopyright{rightsretained}

\acmDOI{}

\acmISBN{}

\acmConference[HRI'21]{ACM HRI conference}{March 2021}{} 
\acmYear{2021}
\copyrightyear{2021}

\begin{document}
	{\sloppy
		\title{Let's be friends! A rapport-building 3D embodied conversational agent for the Human Support Robot}
		
		\author{Katarzyna Pasternak, Zishi Wu, Ubbo Visser}
		\affiliation{%
			\institution{Department of Computer Science\\
				University of Miami}
			\streetaddress{1365 Memorial Drive}
			\city{Coral Gables} 
			\state{FL} 
			\postcode{33146}
		}
		\email{kwp14@miami.edu,{zishi|visser}@cs.miami.edu}
		
		\author{Christine Lisetti}
		\affiliation{%
			\institution{School of Computing and Information Sciences\\
				Florida International University}
			\streetaddress{11200 S.W. 8th Street}
			\city{Miami} 
			\state{FL} 
			\postcode{33199}
		}
		\email{lisetti@cis.fiu.edu}
		
		\renewcommand{\shortauthors}{K. Pasternak et al.}

		\begin{abstract}
			Partial subtle mirroring of nonverbal behaviors during conversations (also known as mimicking or parallel empathy), is essential for rapport building, which in turn is essential for optimal human-human communication outcomes.  Mirroring has been studied in interactions between robots and humans, and in interactions between Embodied Conversational Agents (ECAs) and humans.
			However, very few studies examine interactions between humans and ECAs that are integrated with robots, and none of them examine the effect of mirroring nonverbal behaviors in such interactions. 
			Our research question is whether integrating an ECA able to mirror its interlocutor's facial expressions and head movements (continuously or intermittently) with a human-service robot will improve the user's experience with the support robot that is able to perform useful mobile manipulative tasks (e.g. at home). 
			Our contribution is the complex integration of an expressive ECA, able to track its interlocutor's face, and to mirror his/her facial expressions and head movements in real time, integrated with a human support robot such that the robot and the agent are fully aware of each others', and of the users', nonverbals cues.  
			We also describe a pilot study we conducted towards answering our research question, which shows promising results for our forthcoming larger user study.
			

		\end{abstract}
		
		\maketitle
		
		\section{Introduction}
		
		Forthcoming human-support robots are anticipated to assist people in a variety of contexts \cite{dahl2014robots,vsabanovic2010robots} involving socio-emotional personal information (e.g. helping an elderly person live safely independently) best communicated to humans via their innate communication modalities (e.g. speech, facial expressions, gestures), and ideally with established rapport between interlocutors. We aim to continue investigating the introduction of ECAs on social and service robots as a natural user interface metaphor for HRI \cite{pena2018eeva, goodrich2008human}.
		
		Establishing and maintaining rapport between humans is a proven determinant of positive communication outcomes, and is the result of a combination of highly socio-cultural-emotional complex processes, some of which are unconscious: \textit{mutual attentiveness} (e.g., mutual gaze, mutual 
		interest, and focus during interaction), \textit{positivity} (e.g., head nods, 
		smiles, friendliness, and warmth) and unconscious \textit{coordination} (e.g., postural mirroring, 
		synchronized movements, balance, and harmony) \cite{Grahe1999,Tickle-Degnen1990}.
		
		In this article, we focus on one of these processes, {\em coordination}, of partial subtle mirroring, and synchronized movements of facial expressions, and head movements  \cite{fischer2017mimicking,hess2014emotional,chartrand2005beyond}.  
		Mirroring has been studied extensively in interactions between robots and humans, as well as in interactions between ECAs and humans.
		However, there are only a few studies \cite{Domingo2020gaze, CAVEDON201514} that examine interactions between humans and ECAs that are integrated with robots, and none of them examine the effect of mirroring nonverbal behaviors in such interactions. 
		
		Our aim is to answer the research question as to whether integrating an ECA capable of mirroring its interlocutor's facial expressions and head movements (continuously or intermittently) with a human-service robot will improve the user's experience with the support robot that is capable of performing useful mobile manipulative tasks (e.g. at home). 
		Our current contribution reviews latest research on our topic, and discusses our approach to modeling rapport for human-robot interaction.  We discuss the integration of our speaking, expressive, and realistic ECA with a robotic platform, the Toyota Human-Support Robot (shown in Figure \ref{fig:rapport}), and how we enabled the ECA to sublty track its interlocutor's face, and to mirror their facial expressions and head movements in real time.
		Lastly, we describe a pilot study we conducted towards answering our research question, which shows promising results for our forthcoming larger user study.

		\section{Related work}
		In human-robot interaction (HRI), research on robots establishing rapport is under way, with some research groups investigating  
		verbal  \cite{dieter2019mimic, seo2018investigating, grigore2016talk}, and non-verbal  
		\cite{riek2010my,ritschel2019personalized, hasumoto2020reactive} cues. 
		Previous work has examined how mimicry affects human-ECA interaction and human-robot interaction, but not human interaction with an ECA running on a robot. Hasumoto et al. \cite{hasumoto2020reactive} studied the effects of body movement mimicry in human-robot interaction by designing the Reactive Chameleon, a method of generating robot body movements that subtly mimics human body swaying during interactions. They found that subtle mimicry can positively impact the establishment of rapport, while noticeable mimicry can negatively impact it. However, this method was limited to mimicking movements of the torso and did not consider movement of other parts of the robot such as the head.
		
		In the experiment by Riek et al. \cite{riek2010my}, participants interacted with a robotic chimpanzee named Virgil that exhibited three different types of behavior: full mimicry of head gestures, partial mimicry of nodding gestures only, and no mimicry accompanied by periodic blinks. Afterwards, participants filled out a survey that measured the social attraction toward and emotional credibility of conversation partners. No significant differences were found between participant ratings of the different mimic conditions. However, this might have been due to technical issues, as a few participants ``said that the head movements were too erratic or jerky.'' Other participants wished for the robot to make ``non-speech sounds'' (backchannel cues) to indicate understanding in conjunction with head gestures. This suggests that robot mimicry of nonverbal behavior by itself might not be enough to create rapport during an interaction with a human.
		
		Niewiadomski et al. \cite{Niewiadomski2010} studied how mimicry of smiles influenced interactions between ECAs and humans by testing three different types of ECA smiling behavior when providing backchannel cues: mimicking the smiles of a participant (MS), randomly smiling (RS), and no smiling (NS). They found that participants felt less engaged and more frustrated in condition NS than in condition MS, and ``felt more at ease and more listened to'' while telling a story to the ECA in condition MS than RS. These results suggest that mimicry in the smiling behavior of an ECA influences ``the quality, ease, and warmth, of the user-agent interaction.''

		\begin{figure}
			\includegraphics[width=\columnwidth]{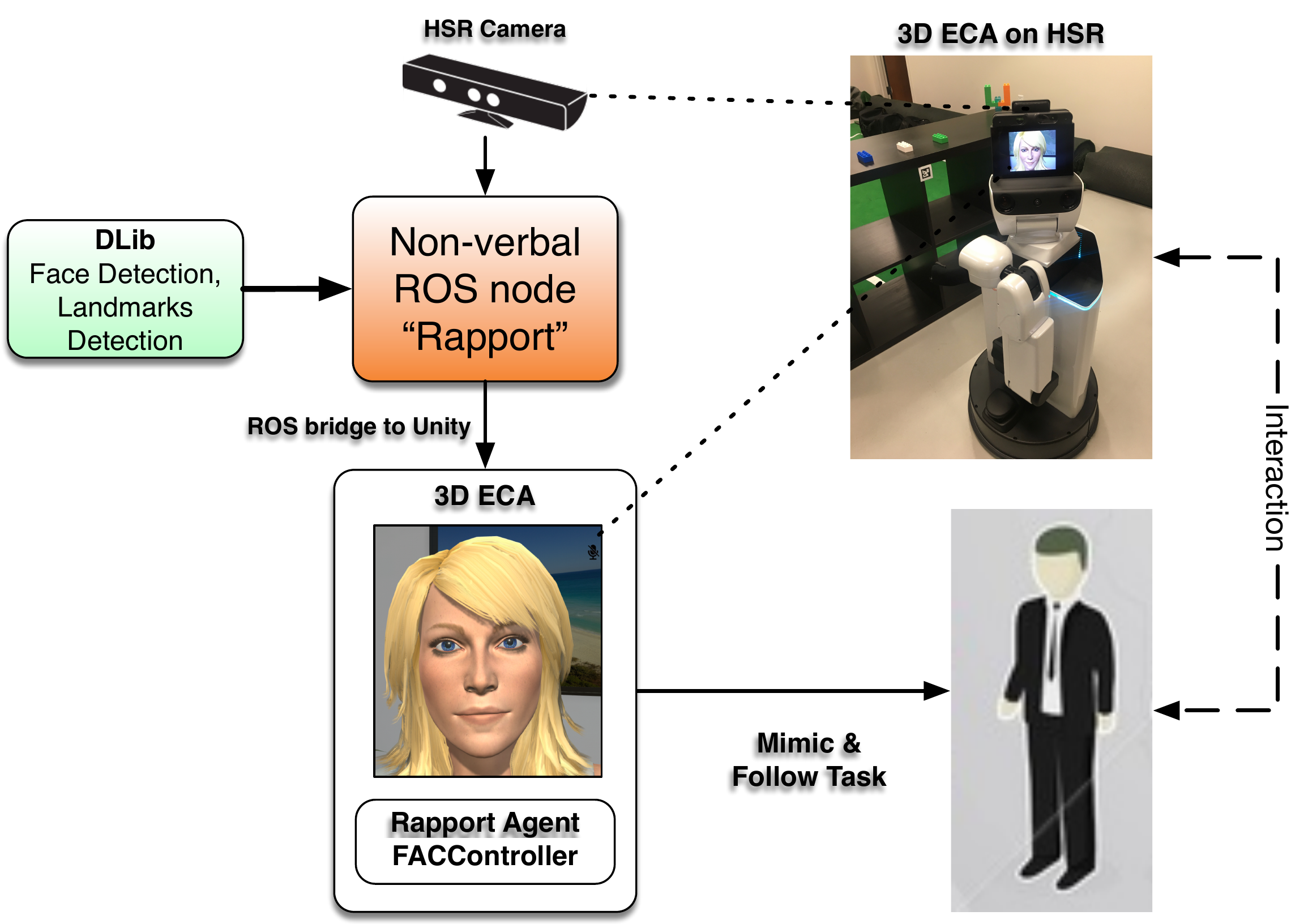}
			\caption{Data flow in our rapport building ROS node.}
			\label{fig:rapport}
		\end{figure}

		In the first experiment of the case study by Stevens et al. \cite{stevens2016mimicry}, participants read some sentences and then listened to an ECA speak some of those sentences incorrectly. They were then asked to say the correct version of the sentence to the ECA. Afterwards, when interacting with the experimental group, the ECA repeated what the subject said while mirroring the eyebrow raises and head nods that were observed during the subject's reading and correcting of potentially erroneous sentences, whereas in the control group no mimicry occurred when the sentence was repeated. The results showed that more prominent cues lead to higher ratings of ECA lifelikeness in the mimic condition, which supports the use of mimicry for building rapport in human-ECA interaction.
		
		
		Although there exists previous work that integrates an avatar on a robot, to the best of our knowledge no study examines how the mimicry of nonverbal behavior by an ECA running on a robot influences interactions with humans.
		For example, Domingo et al. \cite{Domingo2020gaze} projected an avatar on a robotic head and designed a gaze control system that enabled the robotic head to reorient its position based on the location of people that it interacted with. However, the study did not investigate mimicry of nonverbal behavior by the avatar or by the robot.
		
		\section{Approach}
		We propose to integrate the Toyota Human Support Robot (HSR) \cite{Yamamoto2019} with a fully autonomous ECA. HSR is a social robot designed to assist people with disabilities and the elderly with household tasks such as cleaning or bringing objects. HSR has a wide array of sensors that provide rich data on its surrounding environment. By using the Robot Operating System (ROS), which provides services such as access of sensor data from HSR via ROS topics, we can create modules that utilize the sensor data to drive robot behavior during interaction with humans. For an ECA, we use the modular framework eEVA \cite{polceanu2019eeva}, which enables the creation of ECA dialogs suitable for a wide range of scenarios. 
		
		\subsection{Face Detection and Posture Mimicking}
		
		To detect the face of a participant interacting with the HSR, we use an adapted version of DLib \cite{king2009dlib} that gets images from HSR's Asus Xtion Pro RGB-D camera and is able to run with $\approx$ 30fps. When DLib detects a face in the image, it draws a box around the face and marks the face with 68 landmarks as seen in Fig. \ref*{fig:face_det}.
		
		
		Next, we create a ROS node that extracts the central position of participant's face and publishes the extracted data to a ROS topic. This data can then be accessed by both the ECA and HSR for posture mimicry. As the ECA runs on as a stand-alone Unity application, we use the Rosbridge \footnote{\href{http://wiki.ros.org/rosbridge\_suite}{http://wiki.ros.org/rosbridge\_suite}} library to facilitate communication between the ROS node and the ECA.
		
		\begin{figure}
			\includegraphics[width=.8\columnwidth]{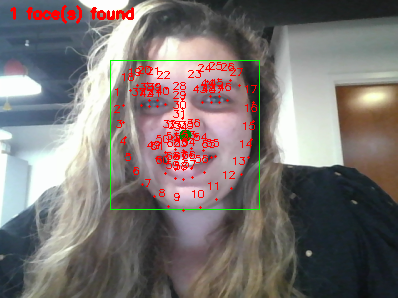}
			\caption{DLib frontal face detection example.}
			\label{fig:face_det}
		\end{figure}

		\subsection{Facial Expression on Emotions Mirroring}
		In addition to mimicking posture, the ability to mirror facial expressions in human-human communication or at least the most universal of such emotive expressions, is important. According to Ekman \cite{Ekman1992a}, there are seven basic emotions that can be expressed by the human face. Our eEVA agent \cite{polceanu2019eeva} (cf. \ref{fig:ekman}) is able to portray any of these emotions in real time through movements of all the individual facial action units identified by Ekman. To enable our system to recognize emotions, we create a ROS node that utilizes EmoPy \cite{angelicaemopy}, a deep learning toolkit that classifies the seven basic emotions from facial expressions (see Fig. \ref{fig:emopy}). 
		We publish the classified emotion from EmoPy to a ROS topic, which is then accessed by eEVA and mapped to facial expressions on the avatar. In that way, the avatar mimics the participant's facial expression during their interaction.
		
		
		\section{Pilot study experiment and Discussion}
		\noindent{\bf Overview.} To answer our research question -- whether integrating an ECA able to mirror its interlocutor's facial expressions and head movements (continuously or intermittently) with a human-service robot will improve the user's experience with the support robot that is able to perform useful mobile manipulative tasks (e.g. at home) -- we designed three within-subjects experiments described in the next sections to assess the impact of various skills on the user's sense of comfort and naturalness with the robot: {\em experiment 1} assesses the impact of posture-mimicking skills; {\em experiment 2} assesses the impact of facial-expression-mirroring skills; and {\em experiment 3} assesses the impact of combining posture-mimicking with facial-expression-mirroring. In preparation for running our planned large user study (with diverse participants, and validated measures of engagement, rapport, and presence), we conducted a pilot study of these experiments and discuss its results.
		
		{\bf Material.}  All three pilot experiments were performed under the following setup: EmoPy and DLib run on an HP Spectre laptop with 16 GB RAM, four CPUs (Intel Core i7-6500U @ 2.50GHz), and an integrated graphics unit (Intel HD Graphics 520, Skylake GT2).
		The ECA was created using the Unity game engine and runs on HSR, which has a CPU board (Intel Core i7-4700EQ @ 2.4GHz) and an NVIDIA Jetson GPU. Communication between the laptop and HSR in close proximity (5 meters) is facilitated using a 5G WiFi network with an average bandwidth speed of 5 ms.
		
		{\bf Participants.}  All three pilot experiments were conducted with three participants recruited from our lab, ages 24 to 56 years.  Participants were 2 (66\%) males and 1 (33\%) female, with mean age of 35 years (SD = 18.2 years).  One  participant reported his race as Asian (33\%), and two reported their race as White (66\%).  Their education level was Graduate Degree (100\%).
		
		{\bf Procedure (in common).}  All three pilot experiments were conducted with the ECA running on the HSR as shown in Fig.~\ref{fig:rapport}.  The HSR was set up in one spot, and lighting conditions remained unchanged.  Depending upon the experiment, participants were asked to take various positions while interacting with the ECA-HSR.
		After interacting with the ECA-robot, participants were asked to answer questions, and an informal discussion followed.  We enabled and disabled modules for each corresponding behavior to control our independent variables.\\
		
		
		\noindent{\bf EXPERIMENT 1: Assessing skills in posture-mimicking}
		In this experiment, we tested the effect on the user of the posture-mimicking skills of the ECA alone, of the robot alone, and of both skills in synchrony.  The module running DLib was used to control the posture mimicking behavior (face following) behavior (Fig.~\ref{fig:face_det}).  Our independent variable was posture-mimicking with three possible conditions: 
		\begin{asparaitem} 
			\item the ECA looks at the person and its face moves according to the user's movements, while the robot stays immobile;  
			\item only the robot head moves following the direction to which the user is moving, while the ECA's face stays immobile in the center of the robot screen; the robot head can eiter move left-right-up-down (all 4 direction and its mixtures), it does not tilt up or down more than $\pm$ 23 degrees, and does not rotate left or right more than $\pm$ 35 degrees); 
			\item both the ECA's face and the robot head move according to the user's movements. 
		\end{asparaitem} 
		
		{\bf Procedure (cont.) for Exp. 1.} 
		Participants were asked to stand in front of the robot (60 cm approx.),  to turn their head to the sides, as well as to move their body, with the restriction that their face should not be turned more than 90 degrees from the camera. This is because DLib can detect faces from front and angled profiles, but not from a complete side profile. Each mode of interaction lasted 15 seconds. 
		At the end of the session, we asked "Which of the three presented versions seemed most comfortable or natural during the interaction? Why?"
		
		\begin{figure}
			\includegraphics[width=.8\columnwidth]{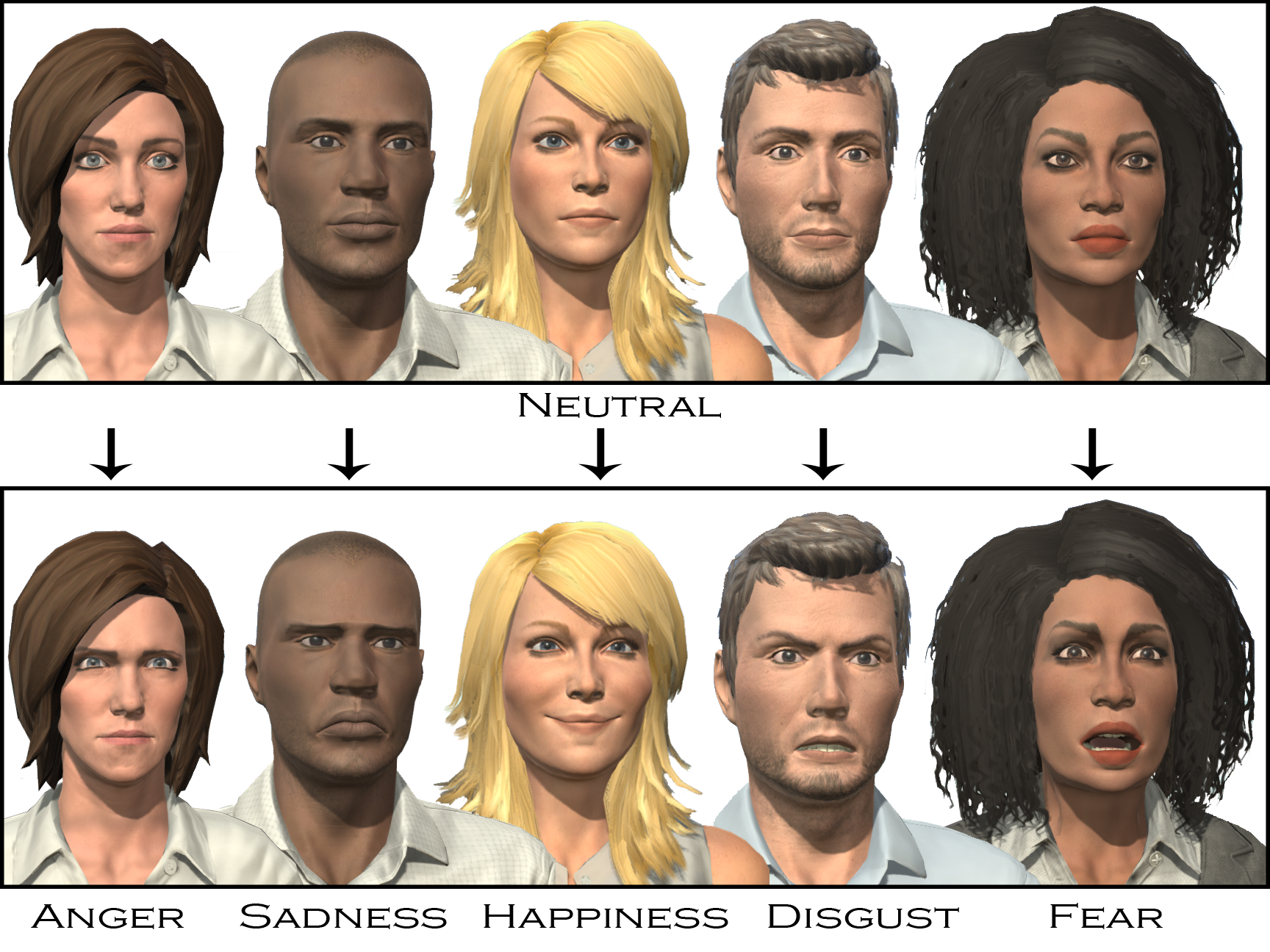}
			\caption{Examples of eEVA facial expressions}
			\label{fig:ekman}
		\end{figure}
		
		{\bf Results.} All participants identified the interaction with the ECA on HSR to be the most natural in the case where only the robot head was moving. They also noted that, in the case where only the ECA moved,  
		it would feel unrealistic if they were walking around the room. The combination of the ECA and the robot head moving in synchrony put the line of sight of the robot away from the participant, which resulted in an impression that the interaction was unnatural. \\
		
		
		\begin{figure}
			\includegraphics[width=.8\columnwidth]{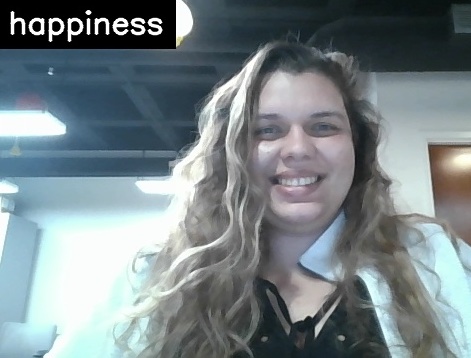}
			\caption{Screenshot of EmoPy's emotion detection.}
			\label{fig:emopy}
		\end{figure}
		
		\noindent{\bf EXPERIMENT 2: Assessing skills in mirroring facial expression of emotions.}
		In this experiment, only the response to the ECA's facial mirroring skills was evaluated.  The module running EmoPy was used to control the emotion mirroring behavior (Fig.~\ref{fig:emopy}). Our independent variable was emotive facial mimicry, with two possible conditions:
		\begin{asparaitem}
			\item emotive facial expression mirroring disabled;
			\item emotive facial expression mirroring active on ECA. 
		\end{asparaitem}
		
		{\bf Procedure (cont.) for Exp. 2.} Each participant were asked to stand in the same proximity to the HSR as before (60 cm), but this time they were asked not to walk around the lab during the experiment. Each person was asked to while looking at the robot, randomly portray three expressions for 10 seconds (to give the system enough time to accomplish the emotion mirroring): happiness, anger, and neutral.  Each mode of interaction lasted 20 seconds.
		Participants stood in front of the HSR's head camera so that their faces would be close enough for expression detection. 
		After the interaction was complete, participants were asked "Which of the two presented versions did you find more engaging? Why?"
		
		{\bf Results.} All participants expressed that, although it was exaggerated and funny at times, the interaction with the ECA on HSR was more engaging when emotive facial expression mirroring was enabled.
		They also noted that sometimes the emotion mirrored by ECA was not the same emotion they were portraying. \\
		
		\noindent{\bf EXPERIMENT 3: Assessing skills combining facial mirroring with posture mimicking.}
		In this experiment, we tested whether posture mimicking, {\em in conjunction with} emotion mirroring, improves user's comfort level during human-robot interaction. Our independent variable was {mimicking combination} with two conditions:
		\begin{asparaitem}
			\item both posture mimicking and emotion mirroring disabled
			\item both posture mimicking and emotion mirroring enabled (the ECA was mirroring emotion and the robot head and ECA were turning towards the location of each participant's face)
		\end{asparaitem}
		
		{\bf Procedure (cont.) for Exp. 3.}
		Participants were asked to walk to the side of the robot while maintaining a close proximity, and to express three different emotions on their face (happiness, anger, neutral) every five to ten seconds, while facing the robot camera. They were given no time restriction on how long they needed to interact with either setup, and asked to give a verbal cue when they wished to end the interaction. When they stopped, they were asked "Which version of the two presented versions (posture mimicking and emotion mirroring both enabled, or neither) did you prefer? Why?"
		
		{\bf Results.} All participants preferred the condition  where both posture mimicking and emotive mirroring were enabled. They explained that the two behaviors made the ECA and the robot more engaging, and helped to establish a connection with the ECA and the robot. \\
		
		
		\noindent{\bf Joint discussion.} During the informal  discussion with all participants, they all agreed that the interaction with the ECA on HSR was more natural with posture mimicry present, while emotive mirroring engaged them more. We also noticed that during the third experiment, participants chose to interact longer when both behaviors were enabled.\\ 
		
		\noindent{\bf Technical considerations.} In all three experiments, we found that the ECA on HSR was performing well under three conditions: (1) the face of a participant was not obstructed; (2) the participant was not outside of the maximal scope of sight of the robot's camera, and (3) the participant was not faced away from the robot. DLib excels at detecting faces in a frontal profile but when a participant turns away, it cannot properly detect their face. 
		The same applies to using EmoPy for detecting emotions on facial expressions. Furthermore, we discovered (1) some latency issues when posture mimicking behavior was enabled (likely due to latency issues in the WiFi connection); and (2) some inaccuracies in the emotions detected by EmoPy (likely due to the small size of the training set used to create models for classifying emotions).
		
		
		\section{Conclusions}
		
		
		To our knowledge, our proposed approach is one of the earliest, if not the first, to study how mimicry of nonverbal behavior by a fully autonomous ECA integrated with a fully autonomous physical service robot influences the agents' interactions with humans. The preliminary experiments conducted on rapport-building -- specifically on posture mimicking and emotive mirroring -- show promising results.
		
		In the first experiment on posture mimicking, participants preferred the robot head moving alone without extra movement from the ECA.
		In the second experiment on emotive mirroring, participants felt more engaged when the ECA mirrored their facial expression, and reported that the interaction felt more natural.
		Finally, in the third experiment that investigated the combined behaviors of posture mimicking and emotive mirroring, participants chose to interact longer when both behaviors were enabled in comparison to when neither were enabled.

		For future work, we will consider alternatives to EmoPy
		to improve the emotive mirroring behavior of our ECA.
		More importantly, we plan to conduct these experiments beyond this pilot with a large enough N and validated measures to produce statistically significant results that can be generalized.
		
		
		
		
		\section{Appendices}
		The video demonstration of the experiments can be found at https://www.cs.miami.edu/home/visser/hsr-videos/nHRI21.mp4.
		\bibliographystyle{ACM-Reference-Format}
		\bibliography{thebibliography,VISAGE} 
	} 
\end{document}